\documentclass[10pt,twocolumn,letterpaper]{article}

\usepackage[pagenumbers]{cvpr} 

\usepackage{graphicx}
\usepackage{amsmath}
\usepackage{amssymb}
\usepackage{booktabs}

\usepackage[accsupp]{axessibility}
\usepackage{acronym}
\acrodef{ID}{In-Distribution}
\acrodef{OoD}{Out-of-Distribution}
\acrodef{ML}{Machine Learning}
\acrodef{FPR}{False Positive Rate}
\acrodef{FNR}{False Negative Rate}
\acrodef{TPR}{True Positive Rate}
\acrodef{TNR}{True Negative Rate}
\acrodef{OC-SVM}{One Class-Support Vector Machine}

\DeclareMathOperator*{\argmax}{argmax} 
\usepackage{mathtools}
\usepackage{calc,bbm,amsthm,bm}

\usepackage{enumitem}

\usepackage{mathtools}

\newcommand{\denselist}{\itemsep 0pt\topsep-6pt\partopsep-6pt}

\usepackage[pagebackref,breaklinks,colorlinks]{hyperref}

\usepackage[capitalize]{cleveref}
\crefname{section}{Sec.}{Secs.}
\Crefname{section}{Section}{Sections}
\Crefname{table}{Table}{Tables}
\crefname{table}{Tab.}{Tabs.}

\def\cvprPaperID{4} 
\def\confName{CVPR}
\def\confYear{2022}

\begin{document}

\title{Class-wise Thresholding for Robust Out-of-Distribution Detection}

\author{Matteo Guarrera \\ 
University of California, Berkeley\\
{\tt\small matteogu@berkeley.edu}
\and
Baihong Jin \\
University of California, Berkeley\\
\and
Tung-Wei Lin \\
University of California, Berkeley\\
\and
Maria A. Zuluaga\\
EURECOM\\
\and
Yuxin Chen\\
University of Chicago\\
\and
Alberto Sangiovanni-Vincentelli\\
University of California, Berkeley
%
%
%
}
\maketitle

\begin{abstract}
We consider the problem of detecting \ac{OoD} input data when using deep neural networks, and we propose a simple yet effective way to improve the \textit{robustness} of several popular \ac{OoD} detection methods against label shift. Our work is motivated by the observation that most existing \ac{OoD} detection algorithms consider all training/test data as a whole, regardless of which class entry each input activates (inter-class differences). Through extensive experimentation, we have found that such practice leads to a detector whose performance is sensitive and vulnerable to label shift. To address this issue, we propose a class-wise thresholding scheme that can apply to most existing \ac{OoD} detection algorithms and can maintain similar \ac{OoD} detection performance even in the presence of label shift in the test distribution.
\end{abstract}

\section{Introduction}
\label{sec:intro}
With the recent advancement in deep learning, image classification has shown great performance improvement under well-controlled settings where the test data are clean and sampled from the same distribution as the training data. However, the deployment of deep learning models in the real world is still full of unknowns. More often than not, well-trained models can come across \acf{OoD} data that are sampled from a different distribution than the one used for training. For example, objects that do not belong to any of the classes in the training data (\ie, \ac{OoD} inputs) can appear at test time. Faced with \ac{OoD} inputs, deep learning-based classifiers may render unpredictable behaviors and often tend to make \textit{overly confident} decisions \cite{nguyen2015deep}. To address this issue, many previous works~\cite{liu2020energybased,liang2020enhancing,hendrycks2019deep,hendrycks2018baseline,lee2018simple} have been dedicated to detect such \acp{OoD} inputs. Therefore, in safety-critical applications such as healthcare and autonomous driving~\cite{joseph2021open}, the classifier should have the ability to yield control to humans upon coming across such inputs, instead of making incorrect predictions silently.

\begin{figure}[tb]
    \centering
    \includegraphics[width=1.01\columnwidth]{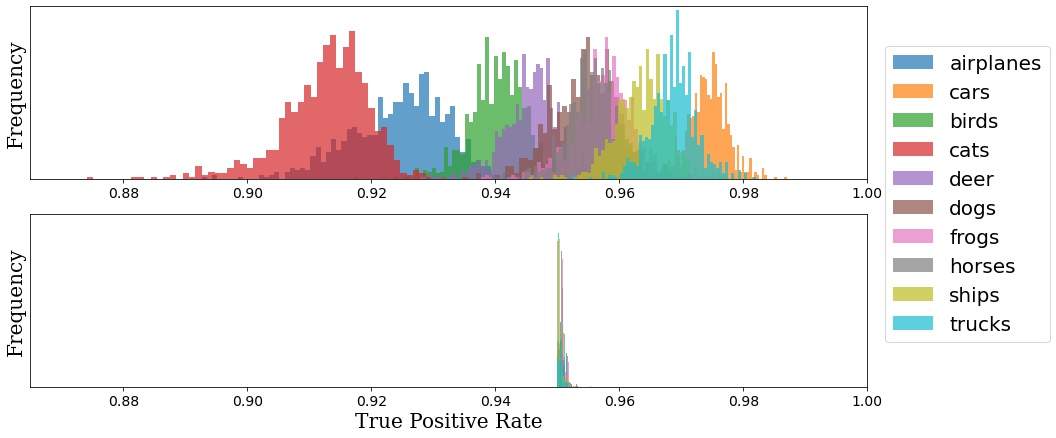}
    \caption{Class-wise True Positive Rate (TPR) variations (\textit{CIFAR10-WideResNet model, max-logit detector}) in simulated scenarios under label shift.
    %
    Narrower histograms in the proposed thresholding scheme (bottom panel), compared to the state-of-the-art (top panel), suggests that our approach is able to guarantee higher robustness through almost constant false alarm rate (5\% in this case).
    %
    %
    }
    \label{fig:tpr-oversample1}
\end{figure}

Among the plethora of works on \ac{OoD} detection, almost all previous literature focuses on improving the detection performance on various \ac{OoD} test sets. Their experiment setups \textit{implicitly} assume that the training and the test \ac{ID} data follow the same distribution
(no distribution shift),
so that the false alarm rates at test time will stay at the same level observed at training. However, it is often not the case in real-world settings, and the distribution shift may result in an increased or decreased number of false alarms on the \ac{ID} data, which can lead to errors incurring into economic losses (additional costs to address these false alarms). Worse still, malicious attackers can exploit this weakness to launch attacks that may cause an overflow of false alarms for certain classes, thus eventually lowering the sensitivity of the detection system against actual \ac{OoD} inputs (due to the excessive number of false alarms injected by attackers). The above-mentioned issues are 
indeed vital for real-world deployment of such systems.

In this paper, we specifically target \ac{OoD} detection algorithms built upon supervised multi-class classifiers, and address the above-mentioned challenges in the context of label shift, a special type of distribution shift, by using a novel thresholding scheme. 
Our approach is applicable even for black-box models, where the internal structure and parameters of the classifier are invisible. The contributions of this paper are three-fold:

\begin{itemize}\denselist
    \item We identify a problem that makes many existing \ac{OoD} detection algorithms vulnerable to test-time label shift.
    \item We propose a \textit{simple} yet \textit{effective} thresholding scheme to address the challenge, and show empirically that our solution can be used as a plugin amendment to any existing OoD system with a class-wise score function. \item Using our novel thresholding scheme, we also assess the performance limit of several \textit{learning-based} \ac{OoD} detectors, and compare them with \textit{non-learning-based} ones. The study provides some guidance on how to navigate the design space of \ac{OoD} detection systems.
\end{itemize}

%
\section{Proposed Approach}\label{sec:approach}
\subsection{Problem formulation}
We consider the \textit{\ac{OoD} detection} problem in \textit{supervised multi-class classification} settings; our goal is to identify whether a data point (image) $\bm{x}$ comes from the distribution $\mathcal{D}_\text{in-dist}$ which the development set data are sampled from. Let us denote a given trained classifier by a function $f_\theta$ whose parameters $\theta$ are learned through a training procedure using data sampled from $D_\text{in-dist}$. 

The training dataset $D_\text{train} = \{(\bm{x_i}, y_i)\}$ is a collection of image and (categorical) label pairs sampled from $\mathcal{D}_\text{in-dist}$, where label $y_i$ for image $\bm{x}_i$ takes integer values from set $\{1,2,\ldots,K\}$, each corresponding to one of the $K$ \ac{ID} classes. The trained classifier $f_\theta$ learns to map an image $\bm{x}_i$ to a \textit{logit} vector $\bm\ell\in\mathbb{R}^K$ that eventually produces a probability vector $\bm{p}$ after softmax transformation. Through the training procedure, the classifier parameters $\theta$ are updated so as to minimize a given loss function for $D_\text{train}$, \eg, the cross-entropy loss.

\paragraph{Non-learning-based detectors}
To detect \acp{OoD}, we need to define an \textit{\ac{OoD} score} function $S$ for each input $\bm{x}$ given classifier $f_\theta$ to indicate how likely the given input is \ac{OoD}. Ideally, an \ac{OoD} detector will always assign higher \ac{OoD} scores to \ac{OoD} data than to \ac{ID} data. In one setup, it is assumed that the \ac{OoD} detector can only observe the output logit vector $\bm\ell$ of classifier $f_\theta(\bm{x})$ but not its internal structure~\cite{liu2020energybased, hendrycks2018baseline,hendrycks2019scaling,liang2020enhancing, hsu2020generalized}; in other words, the model is viewed as a \textit{black box}. This assumption allows \ac{OoD} detection algorithms to be applicable on almost all classifiers. In an alternative setup~\cite{lee2018simple}, the \ac{OoD} detector also has access to the internal structures or the hidden feature mappings of a classification model, which can serve as additional information that is potentially useful for \ac{OoD} detection. 

For the former setup where only the logit vector $\bm\ell$ is available for \ac{OoD} detection, two categories of simple statistics on the logit vector $\bm\ell$, the max-logit score (or the very similar max-softmax confidence score)~\cite{hendrycks2018baseline,hendrycks2019scaling} and the energy score~\cite{liu2020energybased}, are commonly used in literature as the \ac{OoD} score function. For the latter setup, Lee \etal~\cite{lee2018simple} utilized the \textit{Mahalanobis} distance for defining the \ac{OoD} score. 

The max-logit and max-softmax approaches are based on the intuition that \ac{OoD} data will result in lower softmax confidence scores $\max_j p(y=j\,\vert\,\bm{x})$ as well as the corresponding $j$th logit entry $\ell_j$. Depending on which specific statistic to use, we can define two very related detection methods: 
\begin{itemize}[leftmargin=1.2em]
    \item the \textit{max-logit} approach that uses $S_\text{max-logit}(\bm{x}; f_\theta)=-\max_j\ell_j$  as the \ac{OoD} score, and
    \item the \textit{max-softmax} approach that uses $S_\text{max-softmax}(\bm{x}; f_\theta)=-\max_j p(y=j\,\vert\,\bm{x})$ as the \ac{OoD} score.
\end{itemize}
Less confident inputs will receive higher \ac{OoD} scores. The two approaches have demonstrated good detection performance in several empirical studies~\cite{hendrycks2018baseline,hendrycks2019scaling,hendrycks2019deep,hendrycks2019robustness}, and have been widely used as a baseline method due to their simplicity.

The \textit{energy-based} approach~\cite{liu2020energybased} as an alternative to max-logit has also shown good performance in prior literature. Instead of using only the maximum logit entry, the energy-based approach defines the energy function (energy-based \ac{OoD} score)
\vspace{-0.1cm}
\begin{equation}
    S_\text{energy}(\bm{x}; f_\theta) = -T\cdot\log\sum_{j=1}^{K}\exp(\ell_j / T)
\end{equation}
%
as the \ac{OoD} score. Here, $\ell_j$ represents the $j$th entry of the logit vector of $\bm{x}$ and $T$ represents the temperature parameter. It can be proven that the (negative) energy function is a smooth approximation of the maximum logit entry, justifying the similarities among the two statistics for \ac{OoD} detection. Mathematically, we have the following relation between the max-logit score and the energy score,  
\begin{equation} \label{energy_approx}
\max_{1\leq j\leq K} \ell_j < -S_\text{energy}(\bm{x}; f_\theta, T) \leq \max_{1\leq j\leq K}\ell_j + T\cdot\log(K).
\end{equation}

Because the above-mentioned statistics for calculating the \ac{OoD} score are pre-defined and no further learning/tuning is needed, we will hereinafter refer to these methods above as \textit{non-learning-based}.

\paragraph{Learning-based detectors}
As already demonstrated by the energy-based approach~\cite{liu2020energybased}, more \ac{OoD} examples can be detected by considering all logit entries, instead of only the maximum one as in max-logit. One may wonder if we can find another candidate function that performs better. In \textit{learning-based} approaches, the \ac{OoD} score function $S$ is learned or fine-tuned from data with known ID and (optionally) OoD labels. With \ac{ML}, the design space for detectors is hugely expanded, which can produce improved results. Different from non-learning-based methods, learning-based methods infer the decision rule by using the data (\ie, the network's responses) instead of a pre-defined rule. 

The actual choice of learning algorithms relies on the availability of labeling information. If only positive-label (\ie, \ac{ID}) data are available, semi-supervised anomaly detection algorithms can be employed to capture the distribution of \ac{ID} data and differentiate them from \ac{OoD} data. Common choices include anomaly/outlier detection models, such as \ac{OC-SVM} and autoencoders. If some negative-label  (\ie,~\ac{OoD}) data are available, then not only can they serve as validation data for tuning the hyperparameters of semi-supervised models such as \ac{OC-SVM}, but they also enable the use of supervised classification models such as fully-connected neural networks for differentiating between \ac{ID} and \ac{OoD} data.

\begin{figure}[tb]
\centering
\resizebox{\columnwidth}{!}{ 

    \includegraphics[width=1.03\linewidth]{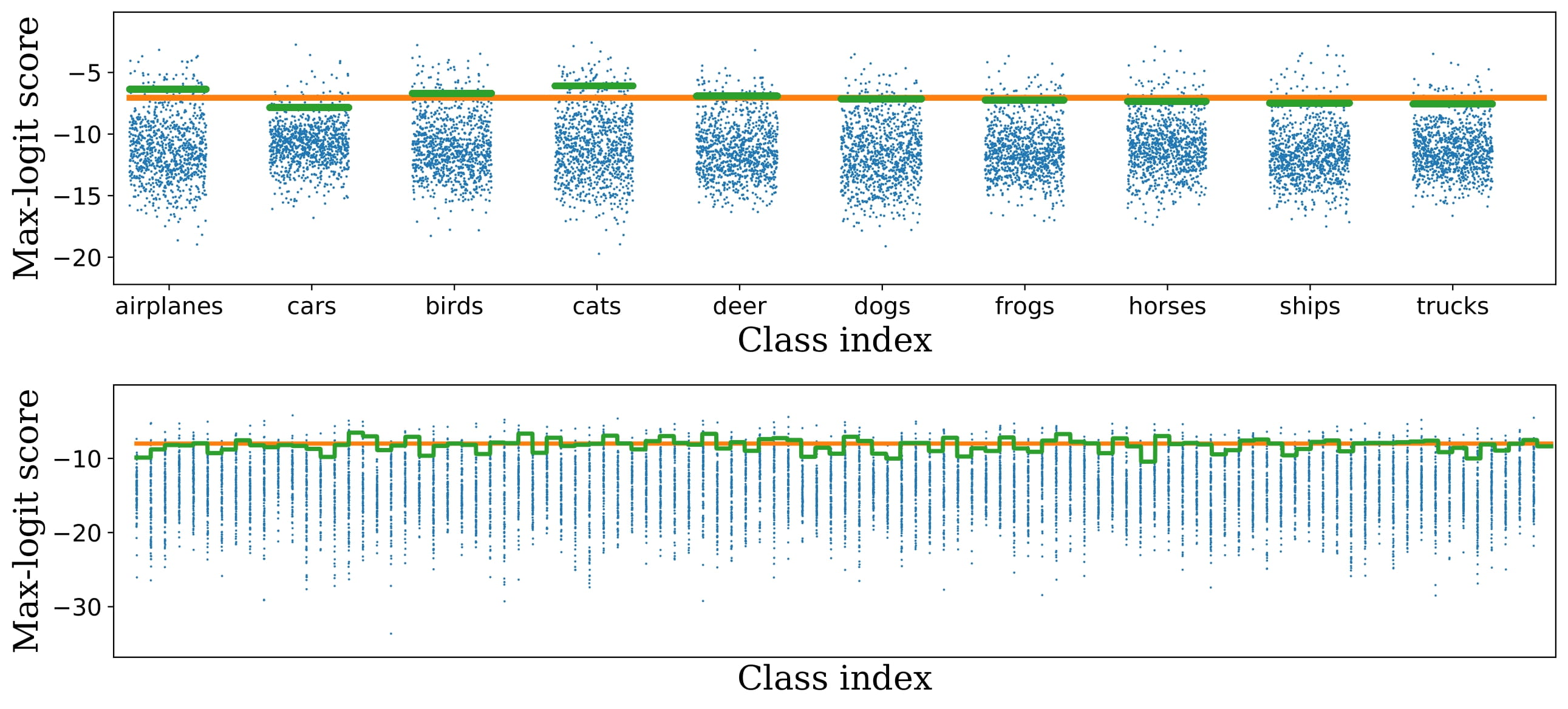}
    }
    
    \caption{For the pre-trained CIFAR10-WideResNet (top panel) and CIFAR100-WideResNet (bottom panel) networks, we display the activated logit entry values (\ie, one that becomes the maximum logit), and also the corresponding \textsc{one} (in orange) and \textsc{multi} (in green) threshold values determined by TPR-95.}
\label{fig:maxscore_thresholds}
\end{figure}

\paragraph{Constant false alarm rate scheme for threshold setting}
Since \ac{OoD} detection is essentially a binary classification problem, a threshold $\tau$ on the \ac{OoD} score $S(\bm{x})$ is needed to dichotomize between \ac{ID} and \ac{OoD}. In mathematical form, we have 
\begin{align}\label{eqn:dichotomy}
    \bm{x}~\text{is}\begin{cases}
    \text{ \textit{out-of-distribution}, if}~S(\bm{x})\geq\tau\\
    \text{ \textit{in-distribution}, otherwise.}
    \end{cases}
\end{align}
Since there is a natural trade-off between false positives (\ie, \ac{OoD} misclassified as \ac{ID}) and false negatives (\ie,  \ac{ID} misclassified as \ac{OoD}) when we modulate the detection threshold $\tau$, a commonly used method for setting $\tau$ is the ``TPR-$\beta$'' thresholding scheme, where $1-\beta\%$ is a preset false alarm rate. In practice, $\tau$ is usually determined on a separate validation set $D_\text{valid}$ (different from $D_\text{train}$ but also sampled from $\mathcal{D}_\text{in-dist}$) to meet the pre-defined false alarm rate level $1-\beta\%$. Although the TPR-$\beta$ method seems to ensure a constant false alarm rate on $\mathcal{D}_\text{in-dist}$, it still suffers from two subtle problems as to be elaborated below.

\paragraph{Non-uniform false alarm rates across in-distribution classes}
Under the above scheme, the false alarm rate for each class may deviate much from the preset level $1-\beta\%$, due to the misalignment among the distributions of each logit entry. To visually illustrate this issue, we repeated the experiments reported in Liu \etal~\cite{liu2020energybased} with the same pre-trained network therein, and plotted in Figure~\ref{fig:maxscore_thresholds} the distribution of the (test-time) logit values for each output node. As we can see, the distributions of the output scores (\ie, the maximum logit values) on each output node are not aligned. A single, unified cutoff threshold (shown in orange) for all the logit entries will result in very different false alarm rates (\ie, the ratios of data points above the set threshold) across the classes.

\begin{figure}[tb]
    \centering
    \includegraphics[width=\columnwidth]{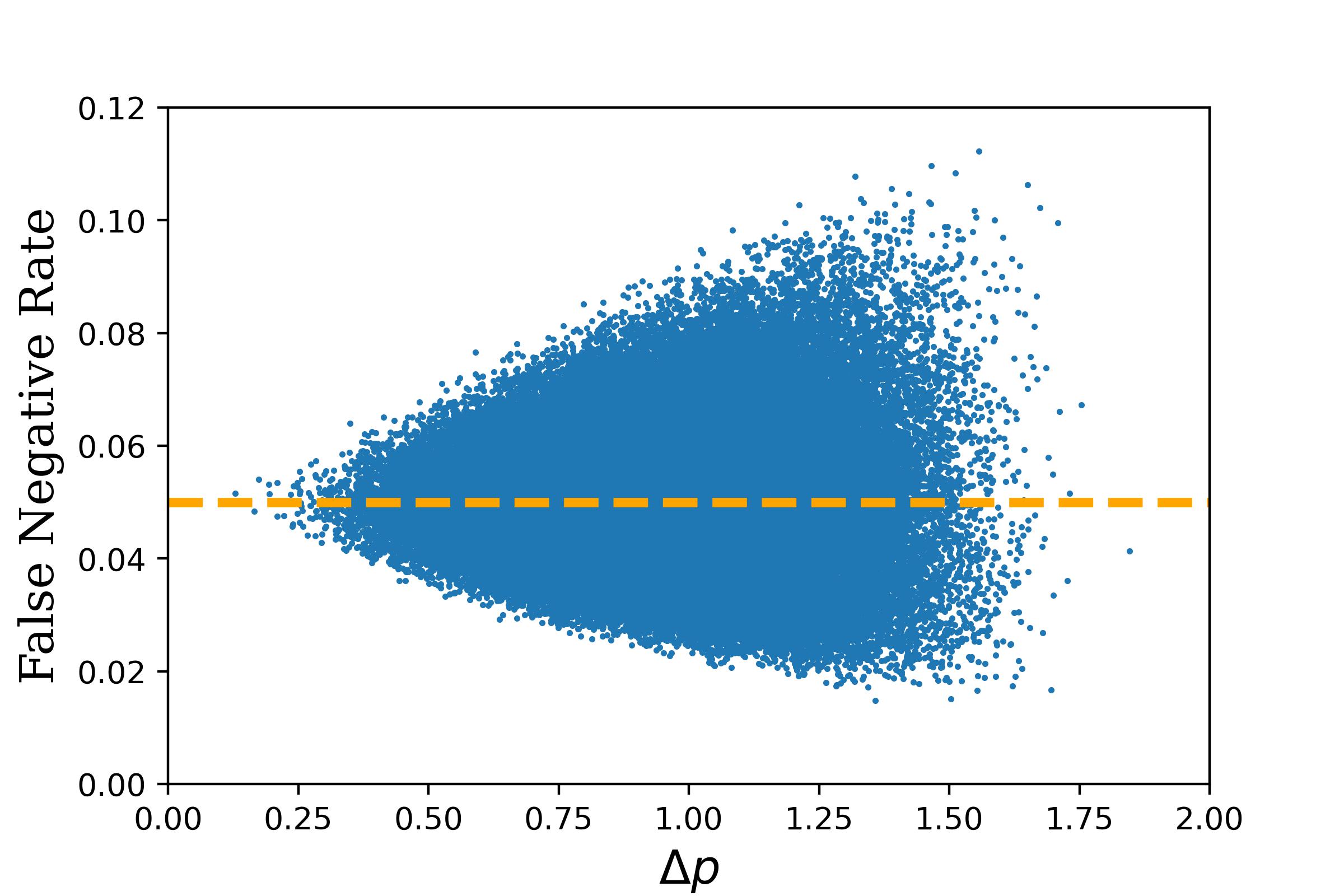}
    \caption{False alarm rate variation under simulated label shift for the pre-trained CIFAR10-WideResNet model. We ran $100000$ simulations and each green point represents one trial.}
    \label{fig:FNR-label-shift}
\end{figure}

\paragraph{Sensitivity to label shift}
In addition to the above issue, the ``constant'' overall false alarm rate can still be fragile, \ie not \textit{robust}, under distribution shift. Particularly, let us take \textit{label shift}~\cite{lipton2018detecting}, a common type of distribution shift, for example. In the presence of label shift, the label marginal $p(y)$ changes but the conditional $p(\bm{x}\,\vert\,y)$ does not. If the label distribution $p(y)$ changes for the test data, the overall false alarm rate can easily fluctuate under the single-threshold approach, due to the varying false alarm rate for each class. To illustrate the issue, we did another experiment on top of the one reported above, again using the outputs of a pre-trained network, to simulate the effect of label shift. Let us denote by $p_\text{train}(y)$ the label marginal of the $\mathcal{D}_\text{in-dist}$ and by $p_\text{test}(y)$ the label marginal of the $\mathcal{D}_\text{in-dist}$, where both $p_\text{train}(y)$ and $p_\text{test}(y)$ are $K$-dimension probability vectors with elements summing up to $1$. In the presence of label shift, $p_\text{train}(y) \neq p_\text{test}(y)$, which as explained above will cause varying false alarm rates under the single-threshold scheme at test time. We performed a simulation study with $10000$ pairs of randomly picked $p_\text{train}(y)$ and $p_\text{test}(y)$ and plotted the resulting false alarm rates against $\Delta{p}\doteq\Vert{p_\text{train}(y)-p_\text{test}(y)}\Vert_2$ (Fig.~\ref{fig:FNR-label-shift}). As can be seen, the spread of resulting false alarm rates at test time becomes larger with increasing $\Delta p$ (\ie, worsening label shift phenomenon), even though we select the threshold hoping to control the false alarm rates to be around $0.05$ (a pre-defined level).

The above observations of fluctuating false alarm rates are concerning, since controlling the number of false alarms is highly important for almost all anomaly/outlier/\ac{OoD} detection tasks. Although the  false alarm rate deviation at test time may be small (say about $1\%$ in the above example), the actual increase in the number of false alarms can be significant since \ac{ID} data (inliers) typically account for the majority of the test data. Worse still, malicious attackers may tamper with the false alarm rates and thereby influence the normal operation of such detection systems. To the best of our knowledge, the above-mentioned issues with existing \ac{OoD} detection algorithms have never been addressed before in \ac{OoD} detection literature. In the upcoming section, we will present a simple yet effective solution to this problem.

\subsection{\textsc{multi}: Class-wise thresholding scheme}\label{sec:multi} 

In the multi-class classification setting, the \ac{ID} data consist of images of multiple classes. Therefore, we can treat $\mathcal{D}_\text{in-dist}$ as a compound distribution of multiple generating processes $\{g_j\}_{j=1}^K$, one for each class. When sampling an image $\bm{x}$ from $\mathcal{D}_\text{in-dist}$, we are actually sampling from the $j$th generating process $g_j$ at probability $p(j)$, \ie, the label marginal for class $j$.

A common assumption in \ac{OoD} literature is that the training data $D_\text{train}$ and the \ac{ID} test data $D_\text{test}$ come from the same distribution $\mathcal{D}_\text{in-dist}$, so that we would expect to get similar false alarm rates on both $D_\text{train}$ and $D_\text{test}$. As we have explained earlier, this assumption may not hold when we deploy the classifier and the \ac{OoD} detector in real-world settings due to issues such as label shift. In the presence of label shift, the label marginals $\{p(j)\}_{j=1}^K$ change but the underlying generating processes $\{g_j\}_{j=1}^K$ stay the same. This suggest us to devise a way to break down the complexity of $\mathcal{D}_\text{in-dist}$ by taking the class label information into account when detecting \acp{OoD}. By considering class-dependent scores, we are designing a dedicated decision rule for the case where the $j$th logit becomes the maximum (\ie, the \textit{activated} logit). We are dividing the entire $K$-dimensional space into $K$ \textbf{disjoint} (non-overlapping) subspaces $\{\mathbb{R}_i^{K}\}_{i=1}^K$, and designing a dedicated decision rule for each subspace. 
Mathematically, let us denote by $\bm{x}^{(j)}$ an input of class $j$, and $\bm{\ell}^{(j)}$ the corresponding logit vector. If $f_\theta$ can correctly classify $\bm{x}^{(j)}$ as class $j$, we have
\begin{align}\label{eqn:new_school}
\bm{\ell}^{(j)} = f_\theta(\bm{x}^{(j)}) \Rightarrow \bm{\ell}^{(j)} \in \mathbb{R}_j^{K} \subset \mathbb{R}^{K}, 
\end{align}
where $\mathbb{R}_j^K = \{\argmax_k \ell_k = j\},~j=1,2,\ldots,K.$
Here, the $\argmax$ operator finds the first occurrence of the maximum entry (in case of multiple occurrences of the same maximum value).

To consider each subspace separately, we can design a separate detection model (\ac{OoD} score function) $S_j$ for each logit subspace $\mathbb{R}_j^K$, or to share the same \ac{OoD} score function $S$ but use different thresholds. 
By setting a different threshold $\tau_j$ for each class, we can treat each predicted class and its subspace as a separate entity. Essentially, \textit{we are enjoying the benefit of having $K$ models, one for each class, in a much less expensive way.}

Based on the above analysis, we can easily extend the single-threshold approach~\eqref{eqn:dichotomy} by using a dedicated threshold $\tau_j$ for each class $j$; this can apply to every detection algorithm that produces a \textit{class-dependent} score, \ie, where each class $j$ is generated by a given $p_j(\bm{x})$. In general, suppose $S(\bm{x})$ is the \ac{OoD} score given by a detection algorithm for input $\bm{x}$, the decision rule under the class-wise thresholding scheme can be written as follows.
\begin{align}\label{eqn:classwise}
    \bm{x}~\text{is}\begin{cases}
    \text{ \textit{out-of-distribution}, if}~S(\bm{x})\geq\tau_j\\
    \text{ \textit{in-distribution}, otherwise}
    \end{cases}
    j=\argmax_k l_k.
\end{align}

\paragraph{Applying the class-wise thresholding scheme to existing \ac{OoD} detection algorithms}
Both non-learning-based and learning-based detectors 
can easily be extended to apply the class-wise thresholding scheme. To be specific, we simply need to replace the single threshold $\tau$ for cutting off the \ac{OoD} score $S(\bm{x};f_\theta)$ with class-wise thresholds $\tau_j$, one for each \textit{activated} logit entry (\ie, the maximum one). The same TPR-$\beta$ scheme can be used to find the $\tau_j$ for each logit entry. In contrast to the previous practice that uses a single, unified cutoff threshold for all logit entries (later referred to as \textsc{one} for brevity), our approach (later referred to as \textsc{multi}) is analogous to a ``switch-case'' statement that uses different thresholds for different activated logits. In our empirical study to be described next, we will compare the performance of \textsc{one} and \textsc{multi} on several popular detection algorithms.


\section{Experiment Results}\label{sec:experiments}

We conducted extensive experiments to compare the performance of \textsc{multi} to that of \textsc{one}. To achieve a fair and comprehensive  comparison between the two thresholding schemes, we performed our experiments using the same or similar settings as in several previous papers. The software implementation can be found as part of the supplementary material.


\paragraph{Datasets} 
We used the same \ac{ID} and \ac{OoD} datasets as in ~\cite{liu2020energybased,lee2018simple} as benchmarks. In addition, to get a more comprehensive view of the performance of the algorithms, we also included another two popular image databases as \ac{ID}, the German Traffic Sign Recognition Benchmark (\textit{GTSRB})~\cite{dset_gtsbr} and ImageNet~\cite{dset_imagenet}. 

For \ac{OoD} benchmarks, we included the ones used in prior works~\cite{liu2020energybased,lee2018simple} in our experiments, including \textit{Places365}~\cite{dset_places} and \textit{SVHN}~\cite{dset_svhn}. In addition, we added a few more from Kaggle and from other works in the literature~\cite{dset_pacs,dset_office} to our evaluation to cover a larger variety of subjects: \textit{Animals}~\cite{dset_animals}, \textit{Anime Faces}~\cite{dset_animeface}, \textit{Fishes}~\cite{dset_fish}, \textit{Fruits}~\cite{dset_fruit}, \textit{iSUN}~\cite{dset_isun}, \textit{Jigsaw Training}~\cite{dset_jigsaw}, \textit{LSUN}~\cite{dset_lsunR_lsunC}, \textit{Office}~\cite{dset_office}, \textit{PACS}~\cite{dset_pacs} and \textit{Texture}~\cite{dset_textures}. 

To reduce the overlap between \ac{ID} and \ac{OoD} datasets, we removed images from \ac{OoD} datasets that share the same class labels as those in \ac{ID} datasets. For example, classes \textit{dog}, \textit{horse}, and \textit{cat} were removed from the \textit{Animals} Dataset since they already existed in CIFAR10, an \ac{ID} dataset in our evaluation.

\paragraph{Pre-trained classification models}
We tested the \ac{OoD} detection algorithms on pre-trained deep learning models of three network architectures: \textit{DenseNet}~\cite{densenet}, \textit{WideResNet}~\cite{zagoruyko2017wide}, and \textit{AlexNet}~\cite{alexnet}. The seven resulting pre-trained models used in our experiments are listed in Table~\ref{tab:max_ene_tpr}. The pre-trained WideResNet models were the same as those used in Liu \etal~\cite{liu2020energybased}; the DenseNet models were from Liang \etal~\cite{liang2020enhancing}. We used the ImageNet-Densenet model from \texttt{torchvision}~\cite{pytorch}, and trained an AlexNet model on the GTSRB dataset.

\begin{figure}[tb]
    \centering
        
    
        \centering
        \includegraphics[width=\columnwidth]{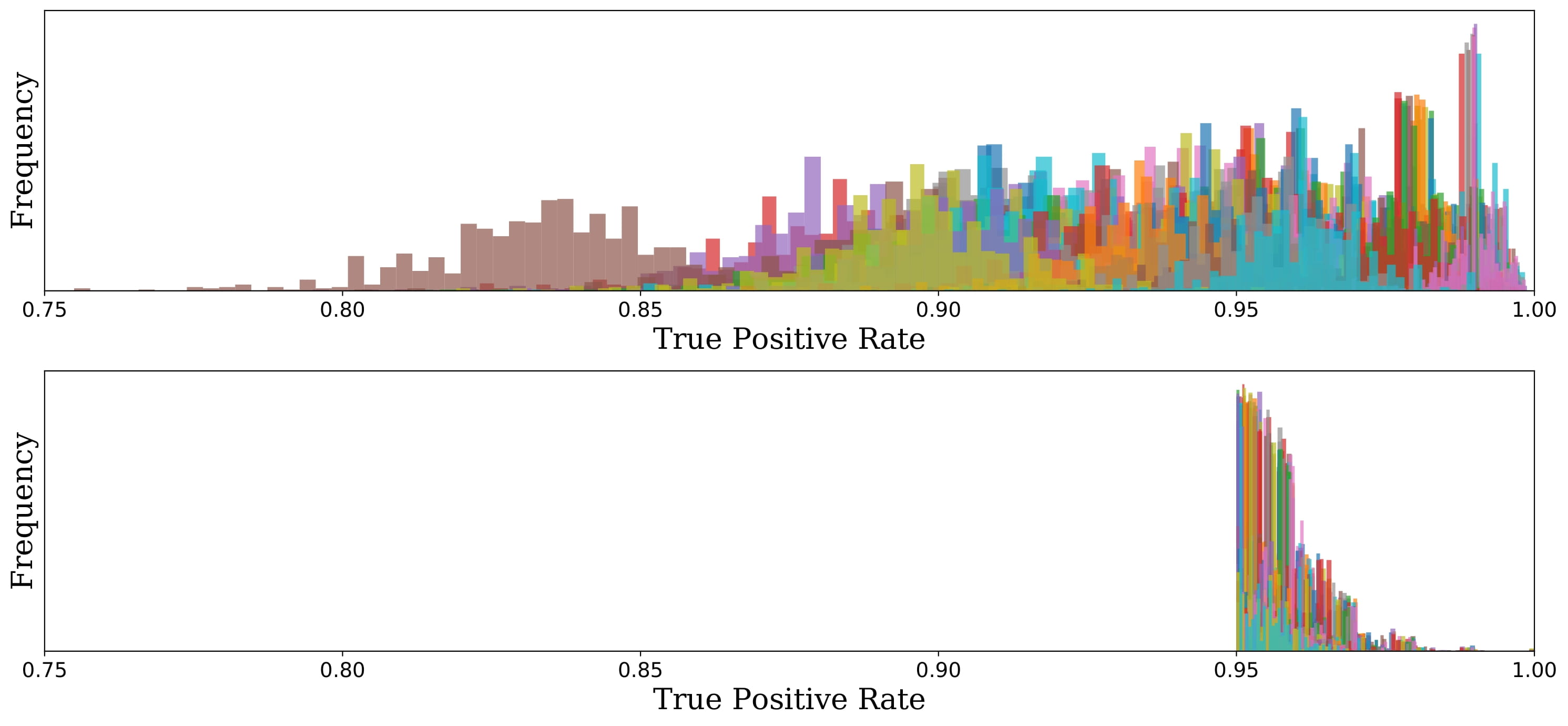}
    \caption{\ac{TPR} variations (as histograms) of each activated logit entry (shown in different colors) of the  CIFAR100-WideResNet model under \textsc{one} (top panel) and \textsc{multi} (bottom panel) for the max-logit detector.}
    \label{fig:tpr-oversample2}

\end{figure}

\paragraph{Setup of \ac{OoD} detectors}
We set up the non-learning-based detectors, max-logit, energy-based and Mahalanobis, using the same methods as described in literature. Then we evaluated the two discussed thresholding schemes, \textsc{one} and \textsc{multi}, by using TPR-95 to determine the respective detection thresholds.

To compare \textsc{one} and \textsc{multi}, we evaluated two anomaly detection models, $k$-NN and \ac{OC-SVM} for learning-based \ac{OoD} detection. In addition, we also tested ODIN~\cite{liang2020enhancing} where an optimal temperature parameter $T$ needs to be learned. As mentioned earlier, learning-based \ac{OoD} detectors have the potential to perform better because of their expressiveness in capturing the density distribution of \ac{ID} data. However, we also noticed the unsatisfactory outcomes from learning-based approaches reported by several previous works. We believe the reported results do not reflect the true potential of learning-based detectors, as it is well-known that hyperparameter settings have profound impacts on the performance of \ac{ML} models. To gauge the full potential of learning-based methods, in our experiments we assessed the the performance limits of learning-based detectors by conducting a \textit{best-case} analysis to measure their \textit{maximum achievable} performance. To do so, we assumed that we have access to a validation set drawn i.i.d. from the test \ac{ID} and \ac{OoD} distribution for tuning the hyperparameters of learning-based detectors.
For each detector, a grid-search was performed over the tunable hyperparameters to select the model instance that achieved the best performance on the test set. This analysis helped us understand the performance limits and the potential room for improvement of learning-based \ac{OoD} detectors.

\begin{table*}[tb]
	\centering
	\caption{\ac{TPR} (unit: \%) of \ac{OoD} Detectors under \textsc{one} and \textsc{multi}.}
	\label{tab:max_ene_tpr}
	\resizebox{2.12\columnwidth}{!}{ 
	    \begin{tabular}{rccccccccccccc}
\toprule
\textbf{} &
  \textit{} &
  \multicolumn{3}{c}{\textit{\begin{tabular}[c]{@{}c@{}}Max-logit Detector\\(\textsc{one}  \(|\)  \textsc{multi})\end{tabular}}} &
  \multicolumn{3}{c}{\textit{\begin{tabular}[c]{@{}c@{}}Energy-based Detector\\(\textsc{one}  \(|\)  \textsc{multi})\end{tabular}}} &
  \multicolumn{3}{c}{\textit{\begin{tabular}[c]{@{}c@{}}k-NN Detector\\(\textsc{one}  \(|\)  \textsc{multi})\end{tabular}}} &
  \multicolumn{3}{c}{\textit{\begin{tabular}[c]{@{}c@{}}OC-SVM Detector\\(\textsc{one}  \(|\)  \textsc{multi})\end{tabular}}} \\
\textbf{Classifier} &
  $K$ &
  \textbf{Min. TPR ($\uparrow$)} &
{  \textbf{Max. TPR ($\downarrow$)}} &
  \textbf{Std. TPR ($\downarrow$)} &
  \textbf{Min. TPR ($\uparrow$)} &
{  \textbf{Max. TPR ($\downarrow$)}} &
  \textbf{Std. TPR ($\downarrow$)} &
  \textbf{Min. TPR ($\uparrow$)} &
{  \textbf{Max. TPR ($\downarrow$)}} &
  \textbf{Std. TPR ($\downarrow$)} &
  \textbf{Min. TPR ($\uparrow$)} &
{  \textbf{Max. TPR ($\downarrow$)}} &
  \textbf{Std. TPR ($\downarrow$)} \\
\midrule

{CIFAR10 - WideResNet} &
  10 &
  91.2  \(|\)  \textbf{94.9} &
  97.5  \(|\)  \textbf{95.0} &
  1.89  \(|\)  \textbf{0.03} &
  91.2  \(|\)  \textbf{94.9} &
  97.6  \(|\)  \textbf{95.0} &
  1.92  \(|\)  \textbf{0.03} &
  88.4  \(|\)  \textbf{94.9} &
  98.8  \(|\)  \textbf{95.0} &
  2.72  \(|\)  \textbf{0.03} &
  90.6  \(|\)  \textbf{94.9} &
  97.8  \(|\)  \textbf{95.0} &
  2.04  \(|\)  \textbf{0.03} \\
{CIFAR10 - DenseNet} &
  10 &
  88.9  \(|\)  \textbf{94.9} &
  98.3  \(|\)  \textbf{95.0} &
  2.52  \(|\)  \textbf{0.03} &
  89.2  \(|\)  \textbf{94.9} &
  98.2  \(|\)  \textbf{95.0} &
  2.45  \(|\)  \textbf{0.03} &
  91.0  \(|\)  \textbf{94.9} &
  98.1  \(|\)  \textbf{95.0} &
  1.85  \(|\)  \textbf{0.03} &
  91.8  \(|\)  \textbf{94.9} &
  97.2  \(|\)  \textbf{95.0} &
  1.49  \(|\)  \textbf{0.03} \\
{SVHN - WideResNet} &
  10 &
  91.0  \(|\)  \textbf{94.9} &
  96.7  \(|\)  \textbf{95.0} &
  1.61  \(|\)  \textbf{0.03} &
  90.8  \(|\)  \textbf{94.9} &
  96.8  \(|\)  \textbf{95.0} &
  1.72  \(|\)  \textbf{0.03} &
  92.4  \(|\)  \textbf{94.9} &
  97.0  \(|\)  \textbf{95.0} &
  1.18  \(|\)  \textbf{0.03} &
  83.4  \(|\)  \textbf{94.9} &
  99.2  \(|\)  \textbf{95.0} &
  4.46  \(|\)  \textbf{0.03} \\
{GTSRB - AlexNet} &
  43 &
  75.9  \(|\)  \textbf{93.3} &
  100.0  \(|\)  \textbf{95.0} &
  5.52  \(|\)  \textbf{0.39} &
  75.0  \(|\)  \textbf{93.3} &
  100.0  \(|\)  \textbf{95.0} &
  5.58  \(|\)  \textbf{0.39} &
  32.5  \(|\)  \textbf{93.3} &
  100.0  \(|\)  \textbf{95.0} &
  13.51  \(|\)  \textbf{0.39} &
  0.8  \(|\)  \textbf{93.3} &
  100.0  \(|\)  \textbf{95.0} &
  22.25  \(|\)  \textbf{0.39} \\
{CIFAR100 - WideResNet} &
  100 &
  83.8  \(|\)  \textbf{93.9} &
  100.0  \(|\)  \textbf{95.0} &
  3.01  \(|\)  \textbf{0.31} &
  83.8  \(|\)  \textbf{93.9} &
  100.0  \(|\)  \textbf{95.0} &
  3.10  \(|\)  \textbf{0.31} &
  72.3  \(|\)  \textbf{93.9} &
  100.0  \(|\)  \textbf{95.0} &
  4.61  \(|\)  \textbf{0.31} &
  84.2  \(|\)  \textbf{93.9} &
  100.0  \(|\)  \textbf{95.0} &
  3.35  \(|\)  \textbf{0.31 }\\
{CIFAR100 - DenseNet} &
  100 &
  82.5  \(|\)  \textbf{93.8} &
  100.0  \(|\)  \textbf{95.0} &
  3.43  \(|\)  \textbf{0.31} &
  81.6  \(|\)  \textbf{93.8} &
  100.0  \(|\)  \textbf{95.0} &
  3.41  \(|\)  \textbf{0.31} &
  76.7  \(|\)  \textbf{93.8} &
  100.0  \(|\)  \textbf{95.0} &
  4.90  \(|\)  \textbf{0.31} &
  78.1  \(|\)  \textbf{93.8} &
  100.0  \(|\)  \textbf{95.0} &
  3.86  \(|\)  \textbf{0.31}\\
  
  ImageNet-DenseNet &  1000 &  72.9  \(|\)  \textbf{90.0} &  100.0  \(|\)  \textbf{94.9} &  4.82  \(|\)  \textbf{0.63}&   69.7  \(|\)  \textbf{90.0} &  100.0  \(|\)  \textbf{94.9} &  4.85  \(|\)  \textbf{0.63} & - &  - & - & - & - & -  \\
  
  \bottomrule

\end{tabular}

    }
\end{table*}

\subsection{False Alarms (Type-1 Errors/False Negatives)}

As described earlier, the single-threshold approach \textsc{one} can result in unevenly distributed false alarms. In our empirical evaluation, we tested both \textsc{one} and \textsc{multi}, and reported in Table~\ref{tab:max_ene_tpr} the \ac{TPR} (one minus the false alarm rate) variation across \ac{ID} classes for all seven pre-trained models. As we can see from the statistics, the baseline scheme \textsc{one} resulted in large performance variation. In several cases, the \ac{TPR} can be as low as $70\%$ under \textsc{one}. On the other hand, \textsc{multi} does not suffer from such problem. The results suggest that, compared to \textsc{one}, \textsc{multi} is much more desirable and robust due to its stable \ac{TPR} performance.

Figure~\ref{fig:tpr-oversample1} and Figure~\ref{fig:tpr-oversample2} highlight the problem associated with \textsc{one} from a different perspective. Here, we modified the number of test-set \ac{ID} data samples for each class by oversampling class $i$ with a random factor $\gamma_i\in[1,10]$, and repeated the same experiment for $1000$ times. The figures show that \textsc{one} not only leads to huge inter-class discrepancies but also induces large intra-class variations among these repeated experiments that simulate different label distribution for the test data. In contrast, \textsc{multi} avoids this problem and gives an almost constant false alarm rate ($5\%$ in this case) despite the label shift.

\begin{figure}[tb]
    \centering
      \begin{subfigure}[t]{\columnwidth}
        \centering
        \includegraphics[width=\columnwidth]{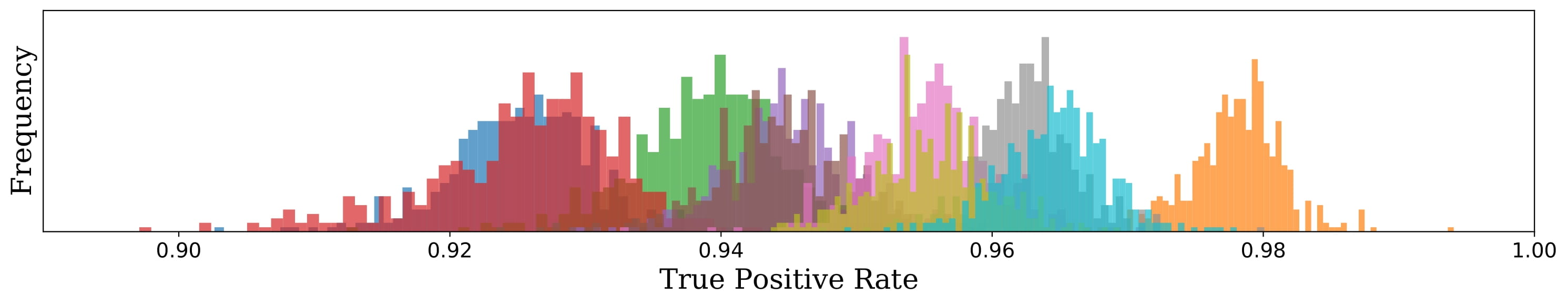}
        
        \caption{\ac{OC-SVM}}
        
        \label{fig:FNR-OCSVM}
      \end{subfigure}
      
      \begin{subfigure}[t]{\columnwidth} 
        \centering

        \includegraphics[width=\columnwidth] {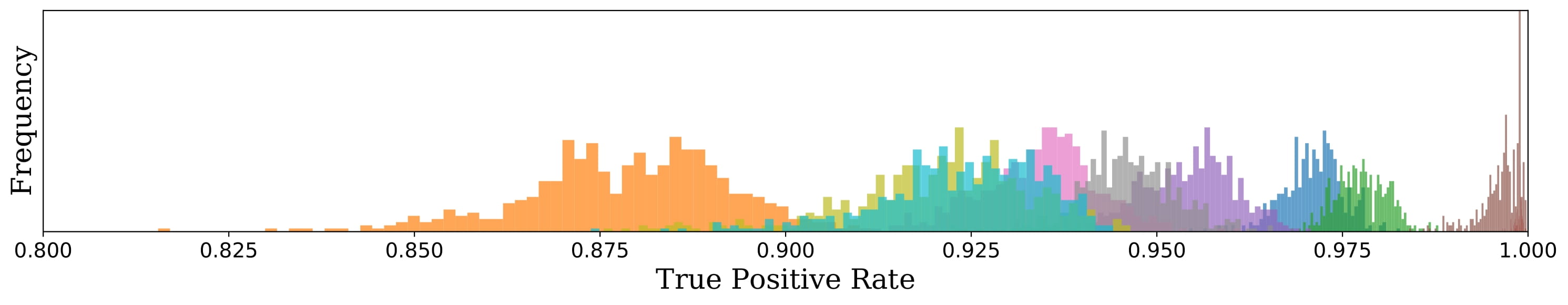} 
        
        \caption{$k$-NN}
        \label{fig:FNR-kNN}
      \end{subfigure}
    \caption{\ac{TPR} variations of each each activated logit entry of CIFAR10-WideResNet model under \textsc{one} for \ac{OC-SVM} (top) and $k$-NN (bottom) detectors.}
    \label{fig:tpr-oversample_max_learning} 
\end{figure}

Figure~\ref{fig:tpr-oversample_max_learning} shows the variations in \ac{TPR} of CIFAR10-WideResNet using \textsc{one} for $k$-NN and \ac{OC-SVM} detectors. Similar to  Figure~\ref{fig:tpr-oversample1} and Figure ~\ref{fig:tpr-oversample2}, the plots indicate large inter-class and intra-class false alarm rate variations, a not so desirable outcome for \ac{OoD} detection applications. This finding again motivates the use of \textsc{multi}.

\paragraph{Simulation Study Shown in Figure~\ref{fig:FNR-label-shift}}
To produce the results shown in Figure~\ref{fig:FNR-label-shift}, we artificially modified the class ratios of the training and test datasets to match randomly chosen class ratios $p_\text{train}(y)$ and $p_\text{test}(y)$. For the CIFAR10-WideResNet case as used in this example, the class ratios $p(y)=(p_1, p_2, \ldots, p_{10})$ are 10-dimension vectors where $\sum_{k=1}^{10} p_k = 1$. In other words, $p(y)\in \mathbb{S}_{10}$ where $\mathbb{S}_{10}$ is a 10-dimension probability simplex. 

Computationally, to randomly sample $p_\text{train}(y)$ and $p_\text{test}(y)$ from $\mathbb{S}_{10}$ , we made use of a property of the exponential distribution $\exp(1)$. Suppose $X_i\sim\exp(1), i=1,2,\ldots,K$ are $K$ i.i.d. samples of an exponential distribution $\exp(1)$. It can be proven that random vector
$$\left(\frac{X_1}{\sum_i X_i},\frac{X_2}{\sum_i X_i},\ldots,\frac{X_K}{\sum_i X_i}\right)$$
is uniformly sampled from $\mathbb{S}_K$. 

By using the trick above, we generated $p_\text{train}(y)$ and $p_\text{test}(y)$, computed the resulting false alarm rate for the test distribution under label shift, and repeated the same experiment for $100000$ times to obtain the scatter plot shown in Figure~\ref{fig:FNR-label-shift}.

\begin{table*}[tb]
	\centering
	\caption{Average Missed Detection Rates under \textsc{one} and \textsc{multi}}
	\label{tab:fpr_full_table}
	\resizebox{0.8\textwidth}{!}{ 
        \begin{tabular}{llccccccc}
\toprule
{} & \textbf{$K$} &      \textbf{Max-softmax} &    \textbf{Max-logit} &       \textbf{Energy} & \textbf{$k$-NN} & \textbf{OC-SVM} & \textbf{ODIN} &  \textbf{Mahalanobis}\\
{} & & \multicolumn{7}{c}{(\textsc{one} \(|\) \textsc{multi})}\\
\midrule
\textbf{CIFAR10-WideResNet } &          10 &  0.57 \(|\) 0.57 &  0.39 \(|\) 0.42 &  0.38 \(|\) 0.41 &  0.35 \(|\) 0.38 & 0.37 \(|\) 0.41 & 0.40 \(|\) 0.40 & 0.45 \(|\) 0.47\\

\textbf{CIFAR10-DenseNet   } &          10 &  0.54 \(|\) 0.54 &  0.39 \(|\) 0.39 &  0.38 \(|\) 0.38 &  0.32 \(|\) 0.34 & 0.35 \(|\) 0.37 & 0.35 \(|\) 0.36 & 0.51 \(|\) 0.52\\

\textbf{SVHN-WideResNet    } &          10 
&  0.09 \(|\) 0.09
&  0.08 \(|\) 0.08 
&  0.08 \(|\) 0.08 
&  0.06 \(|\) 0.06 
& 0.14 \(|\) 0.08 
& 0.17 \(|\) 0.16 & 
0.06 \(|\) 0.04 \\
\textbf{GTSRB-AlexNet      } &          43 &  0.50 \(|\) 0.40 &  0.33 \(|\) 0.35 &  0.32 \(|\) 0.34 &    0.49 \(|\) 0.61 &	0.83 \(|\) 0.66 & 0.25 \(|\) 0.25 &	0.75 \(|\) 0.71\\

\textbf{CIFAR100-WideResNet *} &         100 &  0.79 \(|\) 0.77 &  0.73 \(|\) 0.71 &  0.73 \(|\) 0.71 &    0.77 \(|\) 0.71	 &     0.77 \(|\) 0.76 & 0.67 \(|\) 0.67	 & 0.70 \(|\) 0.67\\

\textbf{CIFAR100-DenseNet  *} &         100 &  0.77 \(|\) 0.75 &  0.72 \(|\) 0.69 &  0.73 \(|\) 0.70 &    	0.81 \(|\) 0.74  &      0.79 \(|\) 0.76 & 0.62 \(|\) 0.62 & 0.69 \(|\) 0.68\\
\textbf{ImageNet-DenseNet  } &        1000 &  0.64 \(|\) 0.70 &  0.57 \(|\) 0.65 &  0.58 \(|\) 0.66 &    - &      - & - & -\\
\bottomrule
\end{tabular}

	}
 	\caption*{\textit{- denotes cases not covered in our experiment due to scalability issues.}\\
 	\textit{* same hyperparameters of the corresponding CIFAR10 classifiers}}
\end{table*}

\subsection{Missed Detections (Type-2 Errors/False Positives)}

Next, let us examine how \textsc{multi} impacts the \ac{OoD} detection performance, in terms of the missed detection rates (ratios of \acp{OoD} that are mistaken as \acp{ID}). We computed the average missed detection rates for the five different \ac{OoD} detection methods over the aforementioned \ac{OoD} benchmarks under both \textsc{one} and \textsc{multi}, for all seven pre-trained classification models. The results are summarized in Table~\ref{tab:fpr_full_table}. 

As we can see from Table~\ref{tab:fpr_full_table}, different pre-trained models yield similar results under \textsc{one} and \textsc{multi}, and we will again take a more detailed look at the results from CIFAR10-WideResNet, shown in Table~\ref{tab:cifar10_wrn_fpr}. As can be seen, the \ac{FPR} performance differences between \textsc{one} and \textsc{multi} are small, up to a few percentage points. Considering the shortcomings of \textsc{one} discussed above, the slight performance trade-off from using \textsc{multi} is well worth it.

In Table~\ref{tab:cifar10_wrn_fpr}, for a given algorithm, we can also see large performance variations across \ac{OoD} benchmark datasets. To get a better understanding of the obtained results, we also calculated the statistical distances between the \ac{ID} training set (\ie, CIFAR10 in this case) and each \ac{OoD} dataset as a measure of their ``OoDness''. Two statistical distance metrics, the \textit{Wasserstein Distance}~\cite{wasserstein_distance} and the \textit{Energy Distance}~\cite{energy_distance}, were used in our calculation. We then computed the Pearson Correlation between the performances of an \ac{OoD} dataset and the corresponding statistical distances (\ie, the correlation between two columns in Table~\ref{tab:cifar10_wrn_fpr}). The results are shown in Table~\ref{tab:fpr_correlation_was_ene}, where we can see negative correlations between the statistical distances and the missed detection rates. This indicates increased difficulties in \ac{OoD} detection for \ac{OoD} datasets that are ``closer'' to the \ac{ID} test set in terms of statistical distances.

It is also worth noticing in Table~\ref{tab:cifar10_wrn_fpr} that the two learning-based detectors ($k$-NN and \ac{OC-SVM}) usually give better performance than the non-learning-based ones (max-logit, energy and Mahalanobis), under both \textsc{one} and \textsc{multi}. Despite the fact that we are conducting a best-case analysis here for learning-based algorithms, this finding still highlights the benefits and potential of learning-based detection algorithms that utilize all logit entries for decision-making.

\subsection{Sensitivity Analysis for Single-Threshold Approach}

In addition to the above-mentioned problems with \textsc{one}, slight variations of $\tau_\text{one}$ can have huge impact on the final \ac{OoD} detection performance at test time. Many factors, including label shift (\ie, varying class ratios in \ac{ID} data), can affect $\tau_\text{one}$. Therefore, using $\tau_\text{one}$ determined on the validation set can yield undesirable detection outcomes. On the contrary, our class-wise thresholding scheme \textsc{multi} is naturally robust to label shift because each class is considered separately. 

\begin{figure}[t]
    \centering
    \includegraphics[width=\columnwidth] {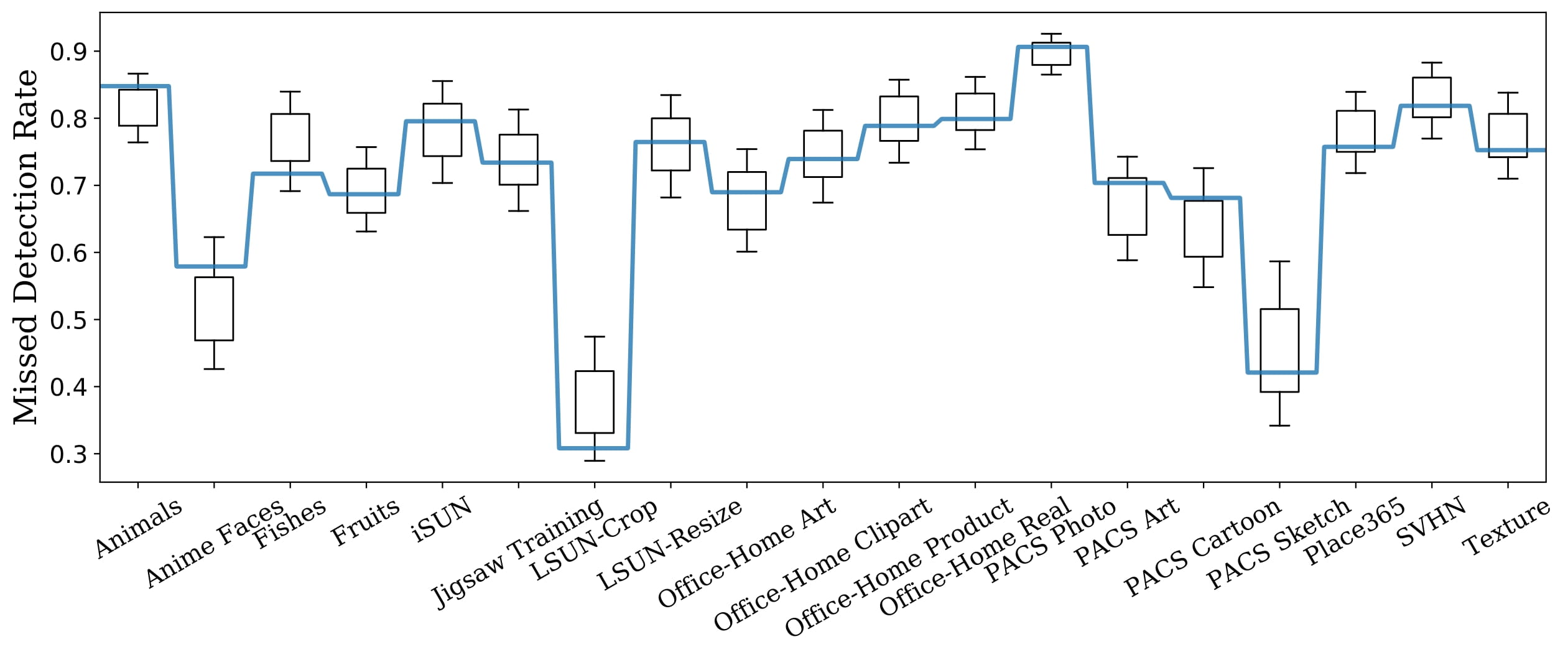}
    
    \caption{Variations of the missed detection rate for \ac{OoD} examples from each benchmark dataset, for perturbed $\tau_\text{one}$ values (box plot). In blue the value for \textsc{MULTI} thresholding scheme.The experiment is performed on pre-trained model CIFAR100-WideResNet, with {max-logit} being used as the \ac{OoD} detection algorithm.} 
    \label{fig:sweeping_th}
\end{figure}

We conducted an experiment to analyze the potential impacts of perturbed $\tau_\text{one}$ due to the choice of the validation set. To simulate additive perturbations $\Delta\tau$ on $\tau_\text{one}$, we uniformly sampled $50$ perturbation values $$\Delta\tau\in\delta\cdot\left[-\left\vert\tau_\text{one} - \min_j\tau_j\right\vert,\left\vert\max_j\tau_j- \tau_\text{one}\right\vert\right],$$
where $\delta\in[0,1]$ is a factor that modulates the maximum strength of the perturbation in the validation set. In our experiments, we set $\delta=0.5$. Figure~\ref{fig:sweeping_th} shows that \textsc{one} often produces worse performance than \textsc{multi} under perturbed threshold $\tau_\text{one}+\Delta\tau$ in terms of missed detection rates, which indicates the risk of using \textsc{one}.

\begin{table*}[tb]
	\centering
	\caption{
	Missed Detection Rates on \ac{OoD} Benchmarks 
	for the CIFAR10-WideResNet Model. }
    \label{tab:cifar10_wrn_fpr}
	\resizebox{\textwidth}{!}{ 
		\begin{tabular}{rccccccccc}
\toprule
\multicolumn{1}{c}{\textbf{}} &
  \textbf{Max-softmax} &
  \textbf{Max-logit} &
  \textbf{Energy} &
  \textbf{k-NN} &
  \textbf{OC-SVM} &
  \textbf{ODIN} &
  \textbf{Mahalanobis} &
  \multicolumn{2}{c}{\textbf{Statistical Distances}} \\
\textbf{Datasets}               & \multicolumn{7}{c}{(\textsc{one} \(|\) \textsc{multi})}                                                     & \textbf{Wasserstein} & \textbf{Energy} \\ \hline
\textbf{Animals}                & 0.68 \(|\) 0.69 & 0.61 \(|\) 0.64 & 0.61 \(|\) 0.64 & 0.63 \(|\) 0.66 & 0.62 \(|\) 0.67 & 0.69 \(|\) 0.69 & 0.80 \(|\) 0.81 & 129                  & 70              \\
\textbf{Anime Faces}            & 0.61 \(|\) 0.68 & 0.38 \(|\) 0.44 & 0.37 \(|\) 0.42 & 0.28 \(|\) 0.36 & 0.32 \(|\) 0.41 & 0.36 \(|\) 0.34 & 0.37 \(|\) 0.45 & 174                  & 95              \\
\textbf{Fishes}                 & 0.51 \(|\) 0.54 & 0.30 \(|\) 0.37 & 0.30 \(|\) 0.36 & 0.28 \(|\) 0.37 & 0.29 \(|\) 0.38 & 0.42 \(|\) 0.42 & 0.30 \(|\) 0.35 & 158                  & 86              \\
\textbf{Fruits}                 & 0.60 \(|\) 0.63 & 0.42 \(|\) 0.48 & 0.42 \(|\) 0.47 & 0.32 \(|\) 0.39 & 0.38 \(|\) 0.47 & 0.55 \(|\) 0.56 & 0.63 \(|\) 0.67 & 181                  & 99              \\
\textbf{iSUN}                   & 0.56 \(|\) 0.51 & 0.35 \(|\) 0.36 & 0.34 \(|\) 0.35 & 0.31 \(|\) 0.32 & 0.33 \(|\) 0.34 & 0.28 \(|\) 0.25 & 0.31 \(|\) 0.27 & 187                  & 102             \\
\textbf{Jigsaw on Training Set} & 0.60 \(|\) 0.58 & 0.45 \(|\) 0.47 & 0.44 \(|\) 0.47 & 0.36 \(|\) 0.39 & 0.43 \(|\) 0.46 & 0.52 \(|\) 0.52 & 0.65 \(|\) 0.64 & 159                  & 88              \\
\textbf{LSUN-Crop}              & 0.31 \(|\) 0.33 & 0.09 \(|\) 0.12 & 0.08 \(|\) 0.11 & 0.11 \(|\) 0.15 & 0.09 \(|\) 0.14 & 0.10 \(|\) 0.10 & 0.31 \(|\) 0.36 & 178                  & 99              \\
\textbf{LSUN-Resize}            & 0.52 \(|\) 0.46 & 0.29 \(|\) 0.30 & 0.28 \(|\) 0.29 & 0.26 \(|\) 0.26 & 0.28 \(|\) 0.29 & 0.22 \(|\) 0.20 & 0.30 \(|\) 0.26 & 193                  & 107             \\
\textbf{Office-Home Art}        & 0.55 \(|\) 0.56 & 0.37 \(|\) 0.40 & 0.36 \(|\) 0.39 & 0.32 \(|\) 0.35 & 0.34 \(|\) 0.39 & 0.40 \(|\) 0.40 & 0.50 \(|\) 0.51 & 151                  & 81              \\
\textbf{Office-Home Clipart}    & 0.55 \(|\) 0.51 & 0.34 \(|\) 0.38 & 0.34 \(|\) 0.38 & 0.34 \(|\) 0.36 & 0.34 \(|\) 0.37 & 0.20 \(|\) 0.22 & 0.12 \(|\) 0.16 & 182                  & 99              \\
\textbf{Office-Home Product}    & 0.60 \(|\) 0.56 & 0.42 \(|\) 0.46 & 0.42 \(|\) 0.45 & 0.44 \(|\) 0.45 & 0.43 \(|\) 0.44 & 0.34 \(|\) 0.36 & 0.37 \(|\) 0.38 & 170                  & 93              \\
\textbf{Office-Home Real}       & 0.57 \(|\) 0.54 & 0.40 \(|\) 0.43 & 0.39 \(|\) 0.42 & 0.37 \(|\) 0.39 & 0.39 \(|\) 0.41 & 0.37 \(|\) 0.38 & 0.41 \(|\) 0.42 & 163                  & 88              \\
\textbf{PACS Photo}             & 0.72 \(|\) 0.71 & 0.62 \(|\) 0.62 & 0.62 \(|\) 0.62 & 0.60 \(|\) 0.62 & 0.61 \(|\) 0.63 & 0.68 \(|\) 0.67 & 0.83 \(|\) 0.82 & 135                  & 75              \\
\textbf{PACS Art}               & 0.58 \(|\) 0.59 & 0.41 \(|\) 0.43 & 0.41 \(|\) 0.43 & 0.36 \(|\) 0.38 & 0.38 \(|\) 0.41 & 0.48 \(|\) 0.47 & 0.59 \(|\) 0.59 & 138                  & 75              \\
\textbf{PACS Cartoon}           & 0.58 \(|\) 0.59 & 0.37 \(|\) 0.40 & 0.37 \(|\) 0.39 & 0.35 \(|\) 0.36 & 0.36 \(|\) 0.40 & 0.31 \(|\) 0.31 & 0.48 \(|\) 0.52 & 150                  & 81              \\
\textbf{PACS Sketch}            & 0.55 \(|\) 0.53 & 0.26 \(|\) 0.30 & 0.25 \(|\) 0.28 & 0.23 \(|\) 0.26 & 0.27 \(|\) 0.30 & 0.18 \(|\) 0.20 & 0.55 \(|\) 0.60 & 178                  & 99              \\
\textbf{Place365}               & 0.59 \(|\) 0.55 & 0.40 \(|\) 0.41 & 0.40 \(|\) 0.41 & 0.38 \(|\) 0.38 & 0.39 \(|\) 0.40 & 0.47 \(|\) 0.47 & 0.69 \(|\) 0.65 & 134                  & 72              \\
\textbf{SVHN}                   & 0.48 \(|\) 0.56 & 0.35 \(|\) 0.41 & 0.35 \(|\) 0.41 & 0.25 \(|\) 0.34 & 0.31 \(|\) 0.37 & 0.44 \(|\) 0.46 & 0.16 \(|\) 0.20 & 214                  & 122             \\
\textbf{Texture}                & 0.60 \(|\) 0.62 & 0.52 \(|\) 0.57 & 0.53 \(|\) 0.57 & 0.44 \(|\) 0.51 & 0.51 \(|\) 0.58 & 0.55 \(|\) 0.54 & 0.17 \(|\) 0.18 & 160                  & 85              \\ \hline
\textbf{Mean Score} &
  0.57 \(|\) 0.57 &
  0.39 \(|\) 0.42 &
  0.38 \(|\) 0.41 &
  0.35 \(|\) 0.38 &
  0.37 \(|\) 0.41 &
  0.40 \(|\) 0.40 &
  0.45 \(|\) 0.47 &
  \multicolumn{1}{l}{} &
  \multicolumn{1}{l}{} \\ \bottomrule
\end{tabular}
	}
	\caption*{\textit{The ``Statistical Distances'' are real-valued scalars measuring the distances between the logits from test-set \ac{ID} and \ac{OoD} examples. More details can be found in the appendix.}}
\end{table*}

\begin{table}[tb]
	\centering
	\caption{Pearson Correlations between the Missed Detection Rates on \ac{OoD} Data and the Corresponding Statistical Distances for CIFAR10-WideResNet Network.}
	\label{tab:fpr_correlation_was_ene}
	\resizebox{0.7\columnwidth}{!}{ 

        \begin{tabular}{rcccc}
\toprule
\multicolumn{1}{l}{} & \multicolumn{2}{c}{\textbf{Wasserstein}} & \multicolumn{2}{c}{\textbf{Energy}}    \\
\textbf{}            & \textsc{ONE}  & \textsc{MULTI}  & \textsc{ONE} & \textsc{MULTI} \\ \hline
\textbf{Max-softmax} & -0.54         & -0.47           & -0.55        & -0.46          \\
\textbf{Max-logit}   & -0.57         & -0.51           & -0.56        & -0.51          \\
\textbf{Energy}      & -0.57         & -0.51           & -0.56        & -0.51          \\
\textbf{$k$-NN}        & -0.66         & -0.60           & -0.65        & -0.59          \\
\textbf{OC-SVM}      & -0.60         & -0.57           & -0.59        & -0.56          \\
\textbf{ODIN}        & -0.57         & -0.56           & -0.55        & -0.54          \\
\textbf{Mahalanobis} & -0.72         & -0.70           & -0.68        & -0.65         \\ \bottomrule
\end{tabular}
	}
\end{table}

\section{Related Work}
\label{sec:releted}
\paragraph{Pre-trained models as black boxes} Hendrycks \etal found out that \ac{OoD} images tend to have lower maximum softmax scores~\cite{hendrycks2018baseline} and maximum logit~\cite{hendrycks2019scaling} than \ac{ID} images. 
Based on the assumption that neural networks are trained to lower the energy of \ac{ID} data, Liu \etal~\cite{liu2020energybased} proposed to use an energy function as a score function. On the other hand, with auxiliary \ac{OoD} datasets and input preprocessing, \textit{ODIN}~\cite{liang2020enhancing} can be considered an \ac{ML} model training the temperature parameter $T$ of the softmax function at the output layer. However, the resulting detector may not generalize well to other unseen \ac{OoD} samples~\cite{hsu2020generalized}. Therefore, Hsu \etal~\cite{hsu2020generalized} proposed \textit{Generalized ODIN} without the use of auxiliary datasets. Still, input preprocessing can induce undesirable time delay for real-time applications.

\paragraph{Pre-trained models with open inner structures} Breaking the black-box assumption, the \textit{Outlier Exposure} (OE) method~\cite{hendrycks2019deep} detects \acp{OoD} by modifying the neural network loss function and with additional training. However, the auxiliary \ac{OoD} datasets that are required are not always accessible~\cite{bergman2020classification}. Lee \etal~\cite{lee2018simple} used \textit{Mahalanobis} distance, and \textit{MOOD}~\cite{lin2021mood} used energy functions, while incorporating information from early layers. 

\paragraph{\ac{OoD} Detection with generative modeling} In addition to the above-mentioned  works that detect \acp{OoD} based on a given classification model, another category of works use a dedicated model for \ac{OoD} detection. An intuitive approach is to use generative modeling to learn the distribution of the \ac{ID} dataset and reject samples that have low likelihood of being \ac{ID}. Although Choi \etal~\cite{choi2018waic} showed that image classifiers can assign higher likelihood to \ac{OoD} data, Ren \etal~\cite{ren2019likelihood} showed how to alleviate this issue by distinguishing between the background and the semantics in the generative model. Zhang \etal~\cite{zhang2020out} explained the phenomenon that flow based models can assign higher likelihood to \ac{OoD} samples, but always generate \ac{ID} images. Tonin \etal~\cite{tonin2021unsupervised} proposed an energy-based unsupervised detection method without access to class labels.


\section{Conclusion}
In this paper, we have addressed an issue that relates to the thresholding strategy used in many state-of-the-art \ac{OoD} detection algorithms. Despite the simplicity of our multi-threshold solution, our main contribution is to identify the problem and to raise people's awareness about the adversarial effect of label shift (and distribution shift in general) in the context of \ac{OoD} detection. This issue, to our best knowledge, has has never been identified or discussed in previous literature. Our solution has addressed this issue in a simple and effective way, making it amenable to real-world applications.

It is worth pointing out one limitation of our proposed approach: when there are very few samples for a particular class in the training and validation sets, it will be difficult to set a meaningful cutoff threshold $\tau_i$. Since real-world training data can often be long-tailed, it is possible that the data for some classes are very scarce. We leave this challenge to our future work. Beyond label shift, we also plan to study how other types of domain shift (such as concept shift and covariate shift) affect \ac{OoD} detection algorithms and how to address such challenges.

\section*{Acknowledgements}

This research was supported by Singapore SinBerBEST project and C3.ai. The first author would like to thank also Fondazione CEUR (Centro Europeo Università e Ricerca) for supporting his research through the program \textit{Giovani Ricercatori}.

{\small
\bibliographystyle{ieee_fullname}
\bibliography{egbib}
}

\newpage
\clearpage
\appendix

\section*{\Large Appendix}

In the appendix, we provide details that are omitted in the main text due to space limitation, which include:
\begin{itemize}\denselist
    \item Details of the experiment setup.
    \item Details of the \ac{OoD} detection algorithms.
    \item An additional experiment to identify important features for differentiating \ac{ID} and \ac{OoD} data.
\end{itemize}

\section{Details of Experiment Setup}

\paragraph{Software and Hardware}
We implemented the deep learning models under the PyTorch framework, and performed training and testing on NVIDIA Tesla V100 GPUs. The other parts of the experiments were done on CPUs. For implementing the \ac{OC-SVM} and $k$-NN anomaly detection algorithms, we utilized \texttt{scikit-learn}~\cite{scikit-learn} and \texttt{PyOD}~\cite{zhao2019pyod} packages; please refer to our attached code for implementation details.

\begin{figure*}[tb]
    \centering
    \resizebox{1.5\columnwidth}{!}{ 
	\includegraphics [width=\textwidth] {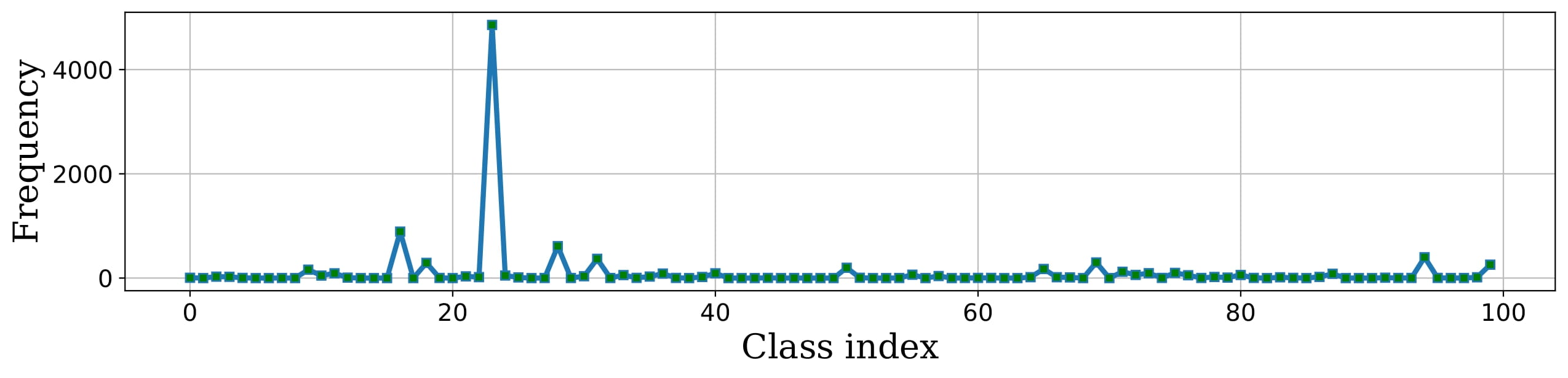} 
	}
    \caption{The uneven distribution of activated logits when the CIFAR100-pre-trained model is fed with the SVHN dataset as \ac{OoD}. The most of images present in the SVHN are indeed classified as a single class, this motivate further the use of a class-wise thresholding scheme, since a given class of \ac{OoD} is most likely to activate a single logit.}
    \label{fig:uneven-activation}
\end{figure*}

\paragraph{\ac{OoD} Benchmarks}
In our experiments, we used \ac{OoD} datasets from a number of different sources as performance benchmarks. A list of these datasets and their sizes can be found in Table~\ref{tab:datasets_size}. Some of them were used by previous literature~\cite{liu2020energybased}. We also used public image datasets such as \textit{Animals-10 Kaggle} Dataset~\cite{dset_animals}. In addition, we created some artificial \ac{OoD} images by modifying \ac{ID} images using \textit{Jigsaw augmentation}~\cite{dset_jigsaw}. In our Jigsaw implementation, each image was divided into $3\times 3$ tiles. Each tile was rotated by multiples of 90 degrees clockwise or counter-clockwise, randomly shuffled, and then recombined into a new image that is considered \ac{OoD}. Furthermore, the \textit{SVHN} Dataset~\cite{dset_jigsaw} was used as \ac{OoD} for \textit{CIFAR10-WideResNet}, \textit{CIFAR100-WideResNet} (Figure \ref{fig:uneven-activation}), \textit{CIFAR10-DenseNet}, \textit{CIFAR100-DenseNet} and \textit{GTSRB-AlexNet} pre-trained models. We standardized both the \ac{ID} and the \ac{OoD} images so that their pixel values would share similar statistical properties.

\paragraph{ODIN Detector}
The ODIN~\cite{liang2020enhancing} detection algorithm uses two strategies, \textit{temperature scaling} and \textit{input pre-processing}, for improving the \ac{OoD} detection performance. Temperature scaling modifies the temperature parameter $T$ in the softmax layer, which adjusts how strong a classifier assumes a closed world as a prior~\cite{hsu2020generalized}. Input pre-processing adds a small controlled noise to each input to make \ac{ID} and \ac{OoD} sample more separable. In our experiment, we chose a noise magnitude $\epsilon=0.0014$ for input pre-processing. 

Despite its effectiveness, ODIN requires \ac{OoD} data to tune hyperparameters for both of temperature scaling and input pre-processing, which leads to two concerns: 1) the \ac{OoD} data may not always be accessible, and 2) the hyperparameters tuned with one \ac{OoD} dataset may not generalize to other \ac{OoD} data. To mitigate the two issues, we followed the Generalized ODIN approach proposed in follow-up work~\cite{hsu2020generalized} and fixed the temperature parameter $T=1000$.

\paragraph{Mahalanobis Detector}
In Lee \etal \cite{lee2018simple} that proposes the Mahalanobis approach, it is assumed that the features from a pre-trained network model follows class-conditional Gaussian distributions. Under this assumption, the confidence score is defined using the Mahalanobis distance with respect to the closest class-conditional distribution, where its parameters are chosen as empirical class means and tied empirical covariance of training samples. As with the ODIN approach~\cite{liang2020enhancing}, we again performed input pre-processing to better separate \ac{ID} and \ac{OoD} samples. We set the noise noise magnitude $\epsilon=0.0014$. 

\paragraph{$k$-NN Detector} $k$-NN~\cite{ramaswamy2000efficient,angiulli2002fast} is a simple, non-parametric supervised classification algorithm; it can also be used in an unsupervised fashion to detect anomalies/outliers/\acp{OoD}. The fundamental assumption of $k$-NN detectors is that the samples from \ac{OoD} inputs tend to stay farther from the clusters of similar observations than \ac{ID} ones. Given a new sample (observation) to classify, its distance to the $k$-th nearest neighbor in the training dataset can be viewed as the \ac{OoD} score.

\begin{table*}[tb]
	\centering
	\caption{Sizes of the \ac{OoD} Datasets Used in our Experiments}
	\label{tab:datasets_size}
    \resizebox{1.5\columnwidth}{!}{
	    \begin{tabular}{rcccc}
\toprule
\textbf{Datasets} &
 
  \textbf{\begin{tabular}[c]{@{}c@{}}CIFAR 10 - 100 \\  ~\cite{dset_cifar} \end{tabular}}

  
  &
  
  \textbf{\begin{tabular}[c]{@{}c@{}}GTSRB ~\cite{dset_gtsbr}\\ German Traffic Sign \\ Recognition Benchmark\end{tabular}} 
  
  
  &
  
  \textbf{\begin{tabular}[c]{@{}c@{}}SVHN ~\cite{dset_svhn}\\ The Street View\\ House Numbers\end{tabular}} 
  
  
  &
   \textbf{\begin{tabular}[c]{@{}c@{}}ImageNet \\  ~\cite{dset_imagenet} \end{tabular}} 
   
  \\ \midrule
\textbf{Test Set}               & 10000 & 12640 & 26032 & 50000 \\ \hline
\textbf{Animals \cite{dset_animals}}                & 10000 & 10000 & 10000 & 10000\\ \hline
\textbf{Anime Faces \cite{dset_animeface}}            & 10000 & 10000 & 10000 & -    \\ \hline
\textbf{Fishes \cite{dset_fish}}                 & 10000 & 10000 & 10000 & -    \\ \hline
\textbf{Fruits \cite{dset_fruit}}                 & 10000 & 10000 & 10000 & -    \\ \hline
\textbf{iSUN \cite{dset_isun}}                   & 8925  & 8925  & 8925  & -    \\ \hline
\textbf{Jigsaw on Training Set \cite{dset_jigsaw}} & 10000 & 10000 & 10000 & -    \\ \hline
\textbf{LSUN-Crop \cite{dset_lsunR_lsunC}}              & 10000 & 10000 & 10000 & -    \\ \hline
\textbf{LSUN-Resize \cite{dset_lsunR_lsunC}}            & 10000 & 10000 & 10000 & -    \\ \hline
\textbf{Office-Home Art \cite{dset_office}}        & 2427  & 2427  & 2427  & 2427 \\ \hline
\textbf{Office-Home Clipart \cite{dset_office}}    & 4365  & 4365  & 4365  & 4365 \\ \hline
\textbf{Office-Home Product \cite{dset_office}}    & 4439  & 4439  & 4439  & -    \\ \hline
\textbf{Office-Home Real \cite{dset_office}}       & 4357  & 4357  & 4357  & -    \\ \hline
\textbf{PACS Photo \cite{dset_pacs}}             & 1282  & 1282  & 1282  & -    \\ \hline
\textbf{PACS Art \cite{dset_pacs}}               & 2048  & 2048  & 2048  & 2048 \\ \hline
\textbf{PACS Cartoon \cite{dset_pacs}}           & 2344  & 2344  & 2344  & 2344 \\ \hline
\textbf{PACS Sketch \cite{dset_pacs}}            & 3929  & 3929  & 3929  & 3929 \\ \hline
\textbf{Place365 \cite{dset_places}}               & 10000 & 10000 & 10000 & -    \\ \hline
\textbf{SVHN \cite{dset_svhn}}                   & 10000 & 10000 & -       & 10000\\ \hline
\textbf{Texture \cite{dset_textures}}                & 5640  & 5640  & 5640  & 5640\\ 

\bottomrule
\end{tabular}
	}
	\caption*{
	\textit{In datasets \textit{Animals-10 Kaggle} and {PACS - Photo Subset}, classes \textit{Cat}, \textit{Dog} and \textit{Horse} have been removed to avoid overlap with the \ac{ID} dataset}}
\end{table*}

\paragraph{\ac{OC-SVM} Detector} \ac{OC-SVM}~\cite{OCSVM} is another commonly used approach for anomaly or outlier detection, and thus can also be used to detect \acp{OoD}. In the \ac{OC-SVM} formulation, a hyperplane is found to separate all data points in the training dataset from the origin. Therefore, given a new sample to classify, its distance from the hyperplane is treated as the \ac{OoD} score.

\vspace{0.2cm}

In order to evaluate the maximum achievable performance, grid search was performed to find the best hyperparameters. We used the $k$-NN module from \texttt{pyOD}~\cite{zhao2019pyod} and the \ac{OC-SVM} module from \texttt{scikit-learn}~\cite{scikit-learn}. from  Table~\ref{tab:hyperparameters_intervals} summarizes the hyperparameters evaluated during grid-search, and Table~\ref{tab:hyperparameters} lists the hyperparameters that were selected and used in producing the results shown in Table~\ref{tab:fpr_full_table}.

\section{Feature Importance Analysis}

We also performed a simple experiment to identify important logit entries to differentiate \ac{ID} and \ac{OoD} samples. Due to the limited space, we did not describe this experiment in the main paper, and deferred it to the appendix.

\begin{figure*}[p]
    \centering
    \includegraphics[width=0.7\textwidth]{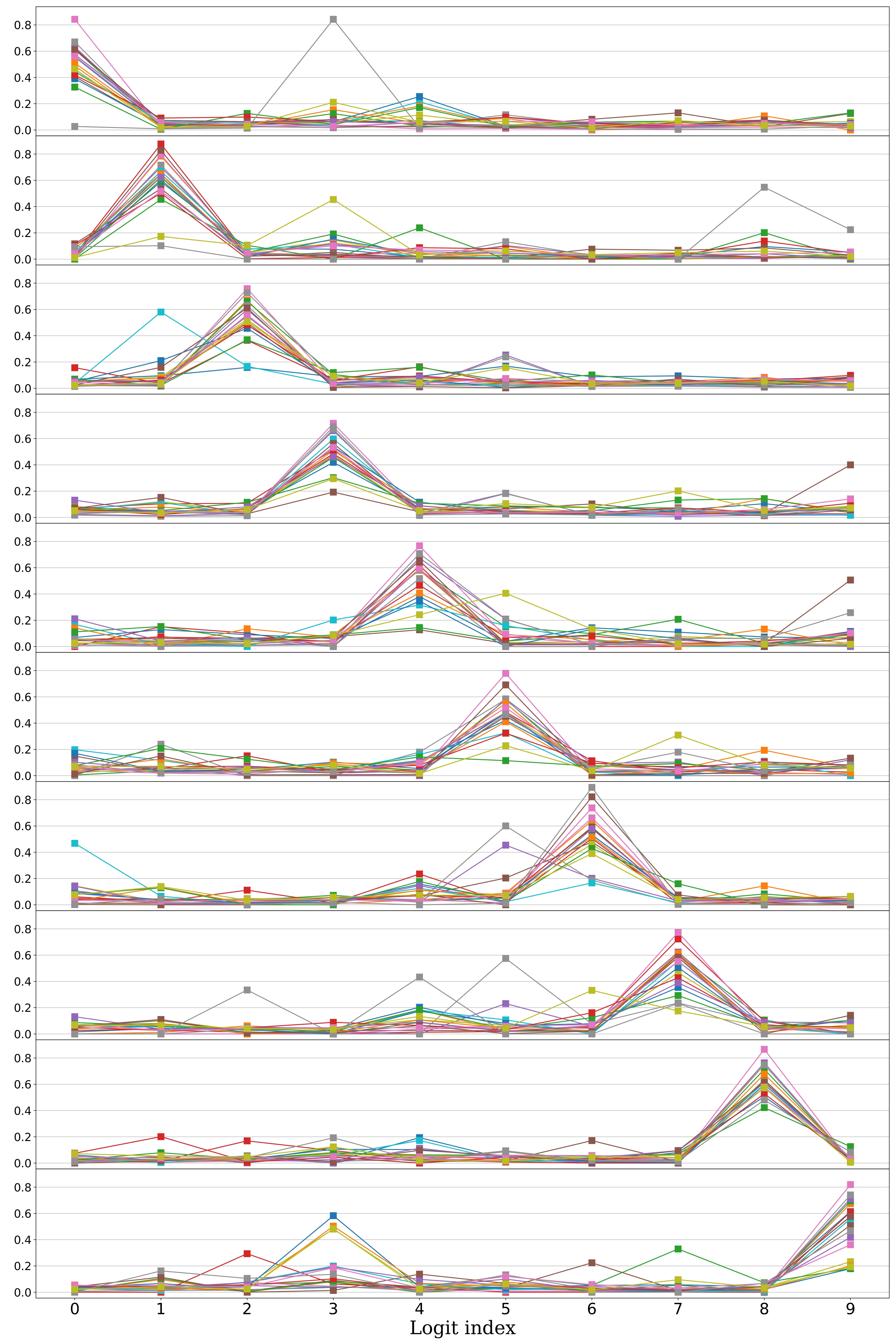}
    \caption{Gini Importance per feature, given a prediction (given argmax), CIFAR10 WideResNet}
    \caption*{\textit{The plot has 10 subplots, one of each represents predicted class, \ie, fixing the argmax. On each subplot on the Y-axis there is the Gini Importance, \ie the importance that that particular feature has during the built of the tree. There are 18 lines in each subplot, for the 18 ood dataset we are considering.}}
    \label{fig:feature-importance}
\end{figure*}

Assuming we have access to the \ac{OoD} test data, we trained a decision tree model (a binary classifier) to differentiate between the \ac{ID} and the \ac{OoD} data, and analyzed the relative importance of each input feature. As we can see from Figure~\ref{fig:feature-importance}, the activated logits (\ie, the diagonal entries in the plot) are almost always assigned the highest importance. This finding indicates that the activated logit entry is a key to differentiating the \ac{ID} and the \ac{OoD} data, which justifies our method to use class-wise decision rules. In addition, we also can notice that some off-diagonal logit entries also receive high feature importance scores, which motivates the use of methods that takes into account all the logit vectors (in energy-based and learning-based methods), instead of solely the maximum one, for improved detection performance.

\begin{table*}[tb]
	\centering
	\caption{Selected Hyperparameters for Learning-based Detectors}
    \label{tab:hyperparameters}
    \resizebox{1.2\columnwidth}{!}{
	\begin{tabular}{lccccccc}
\toprule
        
                            & \multicolumn{3}{c}{\textbf{\ac{OC-SVM}}} & \multicolumn{1}{l}{} & \multicolumn{3}{c}{\textbf{$k$-NN} }         \\ 
                            & kernel   & $\nu$    & $\gamma$   &                      & $k$ & method  & metric     \\ \midrule
WideResNet-CIFAR10          & \textit{poly}     & 0.1   & 1.0     &                      & 4         & \textit{median}  & \textit{braycurtis} \\ 
DenseNet-CIFAR10            & \textit{poly}     & 0.4   & 1.0     &                      & 15        & \textit{largest} & \textit{braycurtis} \\ 
SVHN-WideResNet             & \textit{poly}     & 0.8   & 1.0     &                      & 5         & \textit{mean}    & \textit{braycurtis} \\ 
GTSRB-AlexNet               & \textit{poly}     & 0.8   & 1.0     &                      & 4         & \textit{mean}    & \textit{braycurtis} \\ 
WideResNet-CIFAR100 {[}*{]} & \textit{poly}     & 0.1   & 1.0     &                      & 4         & \textit{median}  & \textit{braycurtis} \\ 
DenseNet-CIFAR100 {[}*{]}   & \textit{poly}     & 0.4   & 1.0     &                      & 15        & \textit{largest} & \textit{braycurtis} \\ 
WideResNet CIFAR10 C        & rbf      & 0.1   & 1.0     &                      & 4         & mean    & \textit{minkowski}  \\
\bottomrule
\end{tabular}
    }
    \caption*{ \textit{* same hyperparameters of the corresponding CIFAR10 classifiers} \newline 
    }
\end{table*}

\begin{table*}[tb]
	\centering
	\caption{Hyperparameters Sweeping Intervals}
    \label{tab:hyperparameters_intervals}
	\resizebox{1.5\columnwidth}{!}{
    \begin{tabular}{cccc}
\toprule

\textbf{$k$-NN}   & \textbf{\#\,Neighbors}                                                                        & \textbf{Metric}                                                                                                   & \textbf{Method}                                                                                                                             \\ \midrule
                & \begin{tabular}[c]{@{}c@{}}4, 5, 6, 7, 8, 9, 10, 11, \\ 12, 13, 14, 15, 20, 25\end{tabular} & \begin{tabular}[c]{@{}c@{}}minkowski, braycurtis, \\ chebyshev, euclidean, manhattan\end{tabular}                 & mean, largest, median       \\ \midrule                          
\textbf{OC-SVM} & \textbf{Kernel}                                                                             & \textbf{$\nu$}                                                                                                       & \textbf{$\gamma$}                                                                                                                              \\ \midrule
                & \begin{tabular}[c]{@{}c@{}}poly, sigmoid,\\ rbf, linear\end{tabular}                        & \begin{tabular}[c]{@{}c@{}}0.01, 0.02, 0.03, 0.04,\\ 0.05, 0.10, 0.15, 0.20,\\ 0.30, 0.40, 0.50, 0.8\end{tabular} & \begin{tabular}[c]{@{}c@{}}0.01, 0.03, 0.05, 0.08, 0.1,\\ 0.2, 0.3, 0.4, 0.5, 0.6, 0.7,\\ 1, 2, 5, 6, 7, 8, 9, 10, scale, auto\end{tabular} 

                \\ \bottomrule
\end{tabular}

	}
\end{table*}

\end{document}